\documentclass{article}

\usepackage[preprint]{neurips_2019}

\usepackage[utf8]{inputenc} 
\usepackage[T1]{fontenc}    
\usepackage{hyperref}       
\usepackage{url}            
\usepackage{booktabs}       
\usepackage{amsfonts}       
\usepackage{nicefrac}       
\usepackage{microtype}      
\usepackage{subfig}
\usepackage[table]{xcolor}
\usepackage{float}
\usepackage{wrapfig}
\usepackage{tikz}
\usepackage{pgfplots}
\pgfplotsset{compat=1.13}
\usepackage{graphicx}
\usepackage{enumitem}
\usepackage{dsfont}
\usepackage{amssymb}
\usepackage{amsmath}
\usepackage{algorithm}
\usepackage[table]{xcolor}

\usepackage{algpseudocode}

\makeatletter
\def\BState{\State\hskip-\ALG@thistlm}
\makeatother

\DeclareMathOperator*{\argmin}{arg\,min}

\makeatletter
\def\BState{\State\hskip-\ALG@thistlm}
\makeatother

\title{Approximating two value functions instead of one: towards characterizing a new family of Deep Reinforcement Learning algorithms}

\author{
  Matthia Sabatelli\\
  Montefiore Institute\\
  Universit\'e de Li\`ege, Belgium \\
  \texttt{m.sabatelli@uliege.be} \\
  \And
  Gilles Louppe \\
  Montefiore Institute\\
  Universit\'e de Li\`ege, Belgium \\
  \texttt{g.louppe@uliege.be} \\
  \And
  Pierre Geurts \\
  Montefiore Institute\\
  Universit\'e de Li\`ege, Belgium \\
  \texttt{p.geurts@uliege.be} \\
  \And
  Marco A. Wiering \\
  Bernoulli Institute for Mathematics, Computer Science \\ and Artificial Intelligence \\
  University of Groningen, The Netherlands \\
  \texttt{m.a.wiering@rug.nl} \\
}

\begin{document}

\maketitle

\begin{abstract}
This paper makes one step forward towards characterizing a new family of \textit{model-free} Deep Reinforcement Learning (DRL) algorithms. The aim of these algorithms is to jointly learn an approximation of the state-value function ($V$), alongside an approximation of the state-action value function ($Q$). Our analysis starts with a thorough study of the Deep Quality-Value Learning (DQV) algorithm, a DRL algorithm which has been shown to outperform popular techniques such as Deep-Q-Learning (DQN) and Double-Deep-Q-Learning (DDQN) \cite{sabatelli2018deep}. Intending to investigate why DQV's learning dynamics allow this algorithm to perform so well, we formulate a set of research questions which help us characterize a new family of DRL algorithms. Among our results, we present some specific cases in which DQV's performance can get harmed and introduce a novel \textit{off-policy} DRL algorithm, called DQV-Max, which can outperform DQV. We then study the behavior of the $V$ and $Q$ functions that are learned by DQV and DQV-Max and show that both algorithms might perform so well on several DRL test-beds because they are less prone to suffer from the overestimation bias of the $Q$ function.

\end{abstract}


\section{Introduction}

Value-based Reinforcement Learning (RL) algorithms aim to learn \textit{value functions} that are either able to estimate how good or bad it is for an agent to be in a particular state, or how good it is for an agent to perform a particular action in a given state. Such functions are respectively denoted as the state-value function $V(s)$, and the state-action value function $Q(s,a)$ \cite{sutton2018reinforcement}. Both can be formally defined by considering the RL setting as a Markov Decision Process (MDP) where the main components are a finite set of states $\mathcal{S}$ $=\{s^{1}, s^{2},...,s^{n}\}$, a finite set of actions $\mathcal{A}$ and a time-counter variable $t$. In each state $s_{t}\in \mathcal{S}$, the RL agent can perform an action $a_{t} \in$ $\mathcal{A}(s_t)$ and transit to the next state as defined by a transition probability distribution $p(s_{t+1} | s_{t}, a_{t})$.
 When moving from $s_t$ to a successor state $s_{t+1}$ the agent receives a reward signal $r_t$ coming from the reward function $\Re (s_{t}, a_{t}, s_{t+1})$. The actions of the agent are selected based on its policy $\pi:\mathcal{S} \rightarrow \mathcal{A}$ that maps each state to a particular action. For every state $s \in \mathcal{S}$, under policy $\pi$ its \textit{value function} $V$ is defined as:
\begin{equation}
    V^{\pi}(s)=\mathds{E}\bigg[\sum_{k=0}^{\infty}\gamma^{k}r_{t+k} \bigg| s_t = s, \pi \bigg],
\end{equation}
which denotes the expected cumulative discounted reward that the agent will get when starting in state $s$ and by following policy $\pi$ thereafter. Similarly, we can also define the \textit{state-action} value function $Q$ for denoting the value of taking action $a$ in state $s$ based on policy $\pi$ as:
\begin{equation}
    Q^{\pi}(s,a)=\mathds{E}\bigg[\sum_{k=0}^{\infty}\gamma^{k}r_{t+k} \bigg| s_t = s, a_t=a, \pi\bigg].
\end{equation}
Both functions are computed with respect to the discount factor $\gamma \in [0,1]$ which controls the trade-off between immediate and long term rewards. The goal of an RL agent is to find a policy $\pi^{*}$ that realizes the optimal expected return:
\begin{align}
 V^{*}(s)=\underset{\pi}{\max}\:V^{\pi}(s), \ \text{for all} \ s\in\mathcal{S}
\end{align}
and the optimal $Q$ value function:
\begin{align}
Q^{*}(s,a)= \underset{\pi}{\max}\:Q^{\pi}(s,a) \ \text{for all} \ s\in\mathcal{S} \ \text{and} \ a \in\mathcal{A}.
\end{align}
It is well-known that optimal value functions satisfy the Bellman optimality equation as given by
\begin{equation}
    V^{*}(s_t) = \underset{a}{\max}\sum_{s_{t+1}}p(s_{t+1} | s_{t}, a_{t}) \bigg[\Re (s_{t}, a_{t}, s_{t+1}) + \gamma V^{*}(s_{t+1}) \bigg]
\end{equation}
for the state-value function, and by
\begin{equation}
    Q^{*}(s_t,a_t)=\sum_{s_{t+1}}p(s_{t+1} | s_{t}, a_{t})  \bigg[\Re (s_{t}, a_{t}, s_{t+1}) + \gamma \: \underset{a_{t+1}}{\max} \: Q^{*}(s_{t+1}, a_{t+1}) \bigg],
\end{equation}
for the state-action value function. Both functions can either be learned via Monte Carlo methods or by Temporal-Difference (TD) learning \cite{sutton1988learning}, with the latter approach being so far the most popular choice among model-free RL algorithms \cite{watkins1992q, rummery1994line, hasselt2010double, pong2018temporal}. In Deep Reinforcement Learning (DRL) the aim is to approximate these value functions with e.g. deep convolutional neural networks \cite{schmidhuber2015deep, lecun2015deep}. This has led to the development of a large set of DRL algorithms \cite{henderson2018deep} among which we mention Deep-Q-Learning (DQN) \cite{mnih2015human} and Double Deep-Q-Learning (DDQN) \cite{van2016deep}. Both algorithms have learning an approximation of the state-action-value function as their main goal. This approach however, has recently shown to be outperformed by the Deep Quality-Value-Learning (DQV) algorithm \cite{sabatelli2018deep}, a relatively novel algorithm which simultaneously approximates the state-value function alongside the state-action value function.

\subsection{The Deep Quality-Value Learning Algorithm}
\label{sec:dqv}

DQV-Learning \cite{sabatelli2018deep} is based on the tabular RL algorithm QV$(\lambda)$ \cite{wiering2005qv}, and learns an approximation of the $V$ function and the $Q$ function with two distinct neural networks that are parametrized by $\Phi$ and $\theta$ respectively. Both neural networks learn via TD-learning and from the same target ($r_{t} + \gamma V(s_{t+1}; \Phi^{-}$)), which is computed by the state-value network $\Phi$. The approximation of the $Q$ function is achieved by minimizing the following loss:
\begin{equation}
    L(\theta) = \mathds{E}_{\langle s_{t},a_{t},r_{t},s_{t+1}\rangle\sim U(D)} \bigg[\big(r_{t} + \gamma V(s_{t+1}; \Phi^{-}) - Q(s_{t}, a_{t}; \theta)\big)^{2}\bigg],
\label{eq:dqv_q_update}
\end{equation}
while the following loss is minimized for learning the $V$ function:
\begin{equation}
L(\Phi) = \mathds{E}_{\langle s_{t},a_{t},r_{t},s_{t+1}\rangle\sim U(D)} \bigg[\big(r_{t} + \gamma V(s_{t+1}; \Phi^{-}) - V(s_{t}; \Phi)\big)^{2}\bigg],
\label{eq:dqv_v_update}
\end{equation}
where $D$ is the Experience-Replay memory buffer \cite{lin1993reinforcement}, used for uniformly sampling batches of RL trajectories $\langle s_{t},a_{t},r_{t},s_{t+1}\rangle$, and $\Phi^{-}$ is the target-network that is used for the construction of the TD errors.
We refer the reader to the supplementary material for a more in-depth explanation of the algorithm which is presented in Algorithm \ref{alg: dqv_algorithm}.

In \cite{sabatelli2018deep} it is shown that DQV is able to learn faster and better than DQN and DDQN on six different RL test-beds. It is however not clear why DQV is able to outperform such algorithms so significantly. In what follows we aim at gaining a more in-depth understanding of this algorithm, and build upon this knowledge to investigate the potential benefits that could come from approximating both the $V$ function and the $Q$ function instead of only the latter one.


\section{Research Questions and Methods}
We now present the different research questions that are tackled in this work. We propose some modifications to the original DQV-Learning algorithm, construct a novel DRL algorithm called Deep-Quality-Value-Max Learning (DQV-Max), and finally investigate the learning dynamics of DQV and DQV-Max by studying their performance under the lens of the \textit{Deadly Triad} phenomenon
\cite{sutton2018reinforcement, van2018deep}.


\subsection{Question 1: does DQV perform better than DQN because it uses twice as many parameters?}

\begin{figure}[ht!]
  \makebox[\textwidth][c]{%
  \subfloat[]{\scalebox{0.25}{\includegraphics{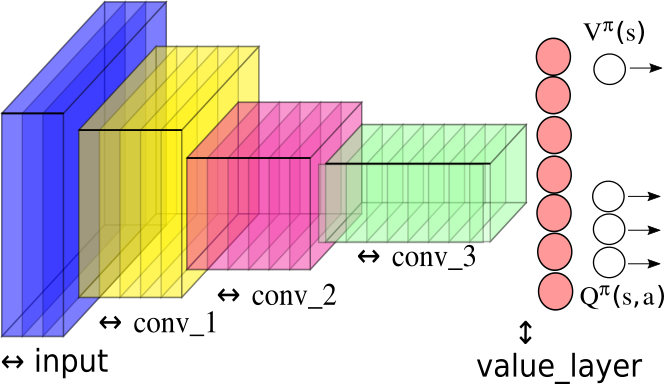}} \label{fig:hard_dqv}}
  \quad
  \subfloat[]{\scalebox{0.25}{\includegraphics{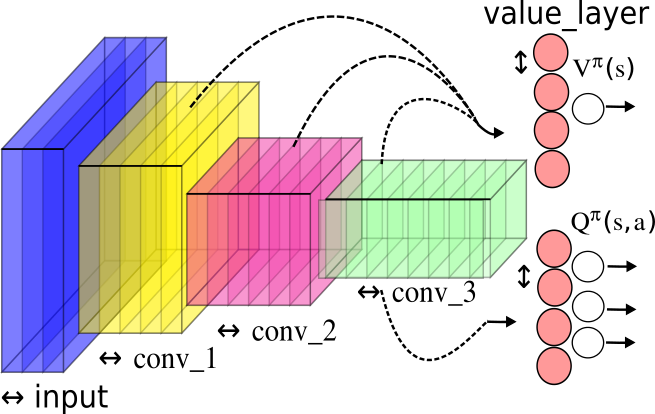}} \label{fig:dueling_dqv}}
  }
\caption{Two representations of convolutional neural networks which jointly approximate the $V$ function and the $Q$ function with the aim of reducing DQV's learning parameters. On the left, an architecture which simply adds one output node to the network next to the output nodes which estimate the $Q$ function. On the right an architecture in which a specific hidden layer precedes the output that is necessary for computing each value function. When it comes to the $V$ function we experiment with different locations of such hidden layer, which is positioned after each possible convolution block.}
\end{figure}

Since the state-value function and the state-action value function are approximated with two separate neural networks, DQV uses twice as many parameters as DQN. It is therefore possible that DQV outperforms this algorithm because the capacity of the model is larger. Approximating two value functions instead of one comes at a price that needs to be paid in terms of memory requirements, a problem which we tackle by taking inspiration from the work proposed in \cite{wang2016dueling} on the \textit{Dueling-Architecture}. Aiming to reduce the amount of trainable parameters of DQV, we modify the original version of the algorithm by exploring two different approaches. The first one, simply adds one output node dedicated for learning the $V$ function next to the nodes that estimate the different state-action values (see Fig. \ref{fig:hard_dqv}). This significantly reduces the parameters of DQV but assumes that the features that are learned for approximating the $V$ function correspond to the ones that are required for learning the $Q$ function. This also assumes that the capacity of the neural network (which structure follows the one proposed in \cite{mnih2015human}) is large enough for approximating both value functions. Since this might not be the case, the second approach adds one specific hidden layer to the network which precedes the outputs of the different estimated value functions (see Fig. \ref{fig:dueling_dqv}). Since, as introduced in \cite{wiering2005qv}, the $V$ function could be easier to learn than the $Q$ function, we also experiment with the location of the hidden layer preceding the $V$ output. We position it after each convolution layer, intending to explore whether the depth of the network influences the quality of the learned $V$
function.

\subsection{Question 2: is using two value function approximators also beneficial in an \textit{off-policy} learning setting?}

One important difference between DQV compared to DQN and DDQN is that the last two algorithms learn the $Q$ function with an \textit{off-policy} learning scheme, while DQV is an \textit{on-policy} learning algorithm. In an \textit{off-policy} learning setting, the TD errors are constructed with respect to values which are different from the agent's actual behavior. This has the benefit of allowing the agent to explore many different policies \cite{van2018deep} because learning is not restricted by the policy that is being followed. If on the one hand, this can be extremely beneficial when it comes to value iteration algorithms \cite{bellman1966dynamic, watkins1992q}, it is also well-known that this particular learning setting yields DRL algorithms that can diverge \cite{sutton2018reinforcement, van2018deep, fujimoto2018addressing, achiam2019towards}. It is therefore not clear whether DQV strongly outperforms DQN and DDQN because it is actually beneficial to approximate both value functions, or simply because being an \textit{on-policy} learning algorithm, DQV is just less prone to diverge.

To answer this question we introduce a novel DRL algorithm called Deep-Quality-Value-Max Learning (DQV-Max). Similarly to DQV this algorithm is also based on a tabular RL-algorithm which was initially introduced in \cite{wiering2009qv}. DQV-Max is constructed in a resembling way as DQV, even though its objectives change. The $V$ function is learned with respect to the greedy strategy $\underset{a\in \mathcal{A}}{\max}\:Q$, therefore making DQV-Max an \textit{off-policy} learning algorithm. DQV's loss that is used for learning the $V$ function presented in Eq. \ref{eq:dqv_v_update} gets modified as follows:
\begin{equation}
L(\Phi) = \mathds{E}_{\langle s_{t},a_{t},r_{t},s_{t+1}\rangle\sim U(D)} \bigg[\big(r_{t} + \gamma \: \underset{a\in \mathcal{A}}{\max}\: Q(s_{t+1}, a; \theta^{-}) - V(s_{t}; \Phi)\big)^{2}\bigg].
\label{eq:dqv_max_v}
\end{equation}

The way the $Q$ function is approximated is not modified, with the only difference being that in this case we decided to not use any target network, since one is already used for learning the $V$ function:
\begin{equation}
    L(\theta) = \mathds{E}_{\langle s_{t},a_{t},r_{t},s_{t+1}\rangle\sim U(D)} \bigg[\big(r_{t} + \gamma V(s_{t+1}; \Phi) - Q(s_{t}, a_{t}; \theta)\big)^{2}\bigg].
    \label{eq:dqv_max_q}
\end{equation}

Similarly as done for DQV, we report an in-depth presentation of this algorithm in Algorithm \ref{alg: dqv_max_algorithm}, which can be found in the supplementary material of this work.


\subsection{Question 3: DQV, DQV-Max, and the \textit{Deadly Triad}, how are they related?}
\label{sec:deadly_triad}

It is known that DRL algorithms are prone to damage the quality of their learned policy when a function approximator that regresses towards itself is used for learning a value function \cite{boyan1995generalization}. This phenomenon is formally known to be caused by the \textit{Deadly Triad} \cite{sutton2018reinforcement}. DQV and DQV-Max are at least connected to two out of three components of the \textit{Triad} and could, therefore, help to gain new insights in the study of DRL divergence. We briefly present how each algorithm is related to each element of the \textit{Triad}, which is highlighted in bold, hereafter:

\begin{itemize}
  \item Both algorithms make obvious use of deep neural networks which serve as \textbf{function approximators} for learning a value function.
  \item DQV and DQV-Max make use of \textbf{bootstrapping}: as shown in Eq.\ref{eq:dqv_q_update} and Eq.\ref{eq:dqv_v_update} DQV bootstraps towards the same target (which is given by the state-value network), whereas DQV-Max bootstraps towards two distinct targets (Eq. \ref{eq:dqv_max_v} and Eq. \ref{eq:dqv_max_q}).
  \item Lastly only DQV-Max presents the final element of the \textit{Triad}: differently from DQV this is the only algorithm which learns in an \textbf{off-policy} learning setting.

\end{itemize}

Based on this information we formulate two hypotheses: the first one is that being an \textit{on-policy} learning algorithm, DQV should be less prone to suffer from divergence than more popular \textit{off-policy} algorithms. The second hypothesis is that even though DQV-Max is an \textit{off-policy} learning algorithm, there is one important difference within it: it learns an approximation of the $Q$ function with respect to TD errors which are given by the $V$ network, therefore not regressing the $Q$ function towards itself anymore (as DQN and DDQN do). Because of this specific learning dynamic, we hypothesize that DQV-Max, could be less prone to diverge. However, if compared to DQV, DQV-Max should still diverge more since it presents all elements of the \textit{Deadly Triad}, while DQV does not.

 To quantitatively assess our hypotheses we will investigate whether DQV and DQV-Max suffer from what is known to be one of the main causes of divergence in DRL \cite{van2018deep}: the overestimation bias of the $Q$ function \cite{hasselt2010double}.


\section{Results}

We now answer the presented research questions by reporting the results that we have obtained on the well-known \texttt{Atari-Arcade-Learning} (ALE) benchmark \cite{bellemare2013arcade}. We have used the games \texttt{Boxing, Pong} and \texttt{Enduro} since they are increasingly complex in terms of difficulty and can be mastered within a reasonable amount of training time. We refer the reader to the supplementary material for a thorough presentation of the experimental setup that has been used in this work.

\subsection{HARD-DQV and Dueling Deep Quality-Value Learning (Question 1)}
\label{sec:shared_dqv}

\begin{figure}[ht!]
\includegraphics[width=.3\textwidth]{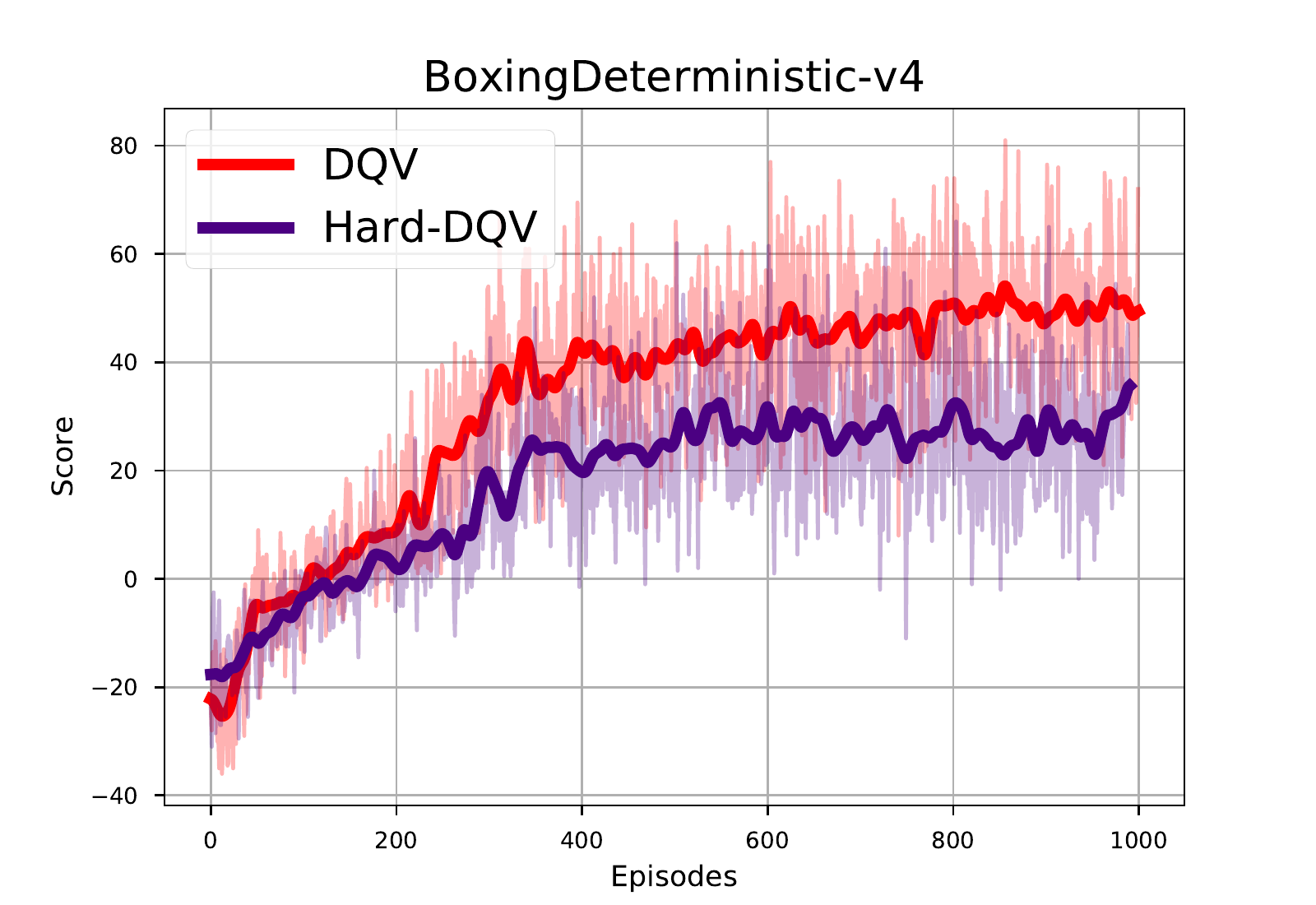}\hfill
\includegraphics[width=.3\textwidth]{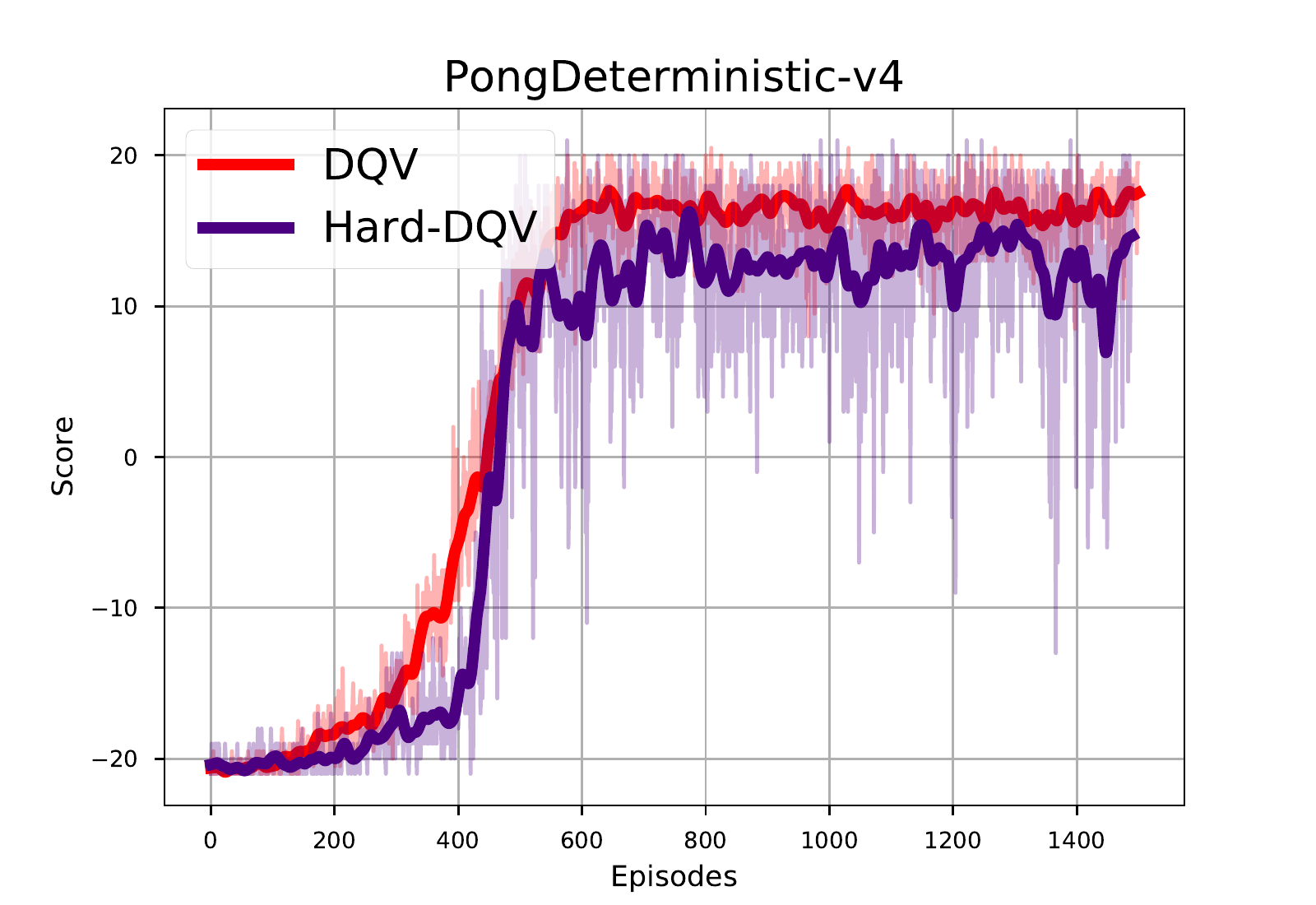}\hfill
\includegraphics[width=.3\textwidth]{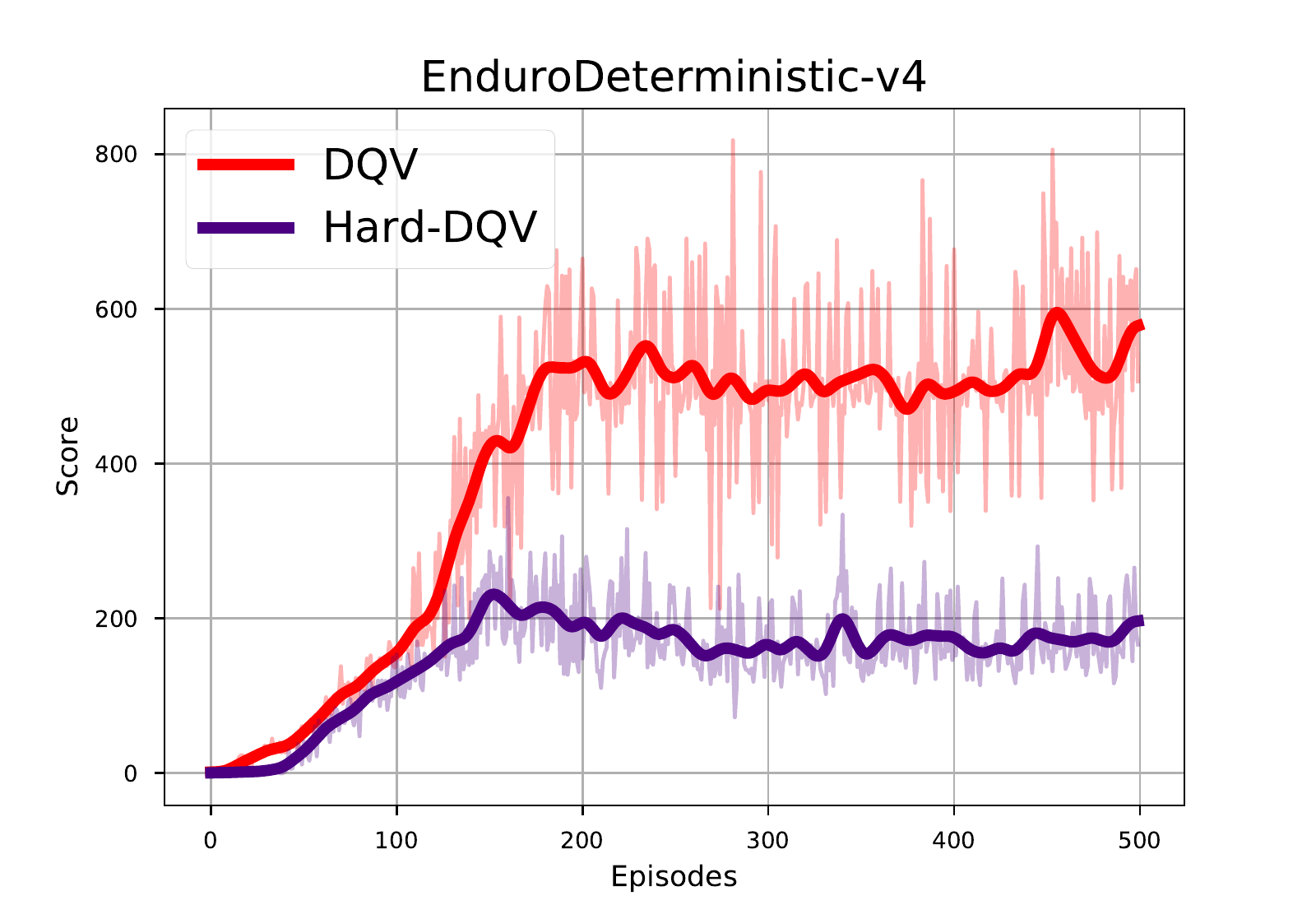}
\caption{The results obtained with a neural architecture in which the output node dedicated to learning the $V$ function, is placed next to the output nodes which estimate the $Q$ function. It is clear from the results that this approach, albeit drastically reducing the memory requirements of DQV, also significantly harms its performance.}
\label{fig:hard-shared-dqv}
\end{figure}

We start by reporting the results that have been obtained with the neural architecture reported in Fig. \ref{fig:hard_dqv}. We refer to this algorithm as HARD-DQV since all the parameters of the agent are hardly-shared within the network \cite{caruana1997multitask}. We can observe from the blue learning curves presented in Fig. \ref{fig:hard-shared-dqv} that this approach drastically reduces the performance of DQV on all the tested environments. The results are far from the original DQV algorithm and suggest that a proper approximation of the $V$ function and $Q$ function can only be learned with a neural architecture which has a higher capacity and that reserves some specific parameters that are selective for learning one value function.

\begin{figure}[ht!]
\includegraphics[width=.3\textwidth]{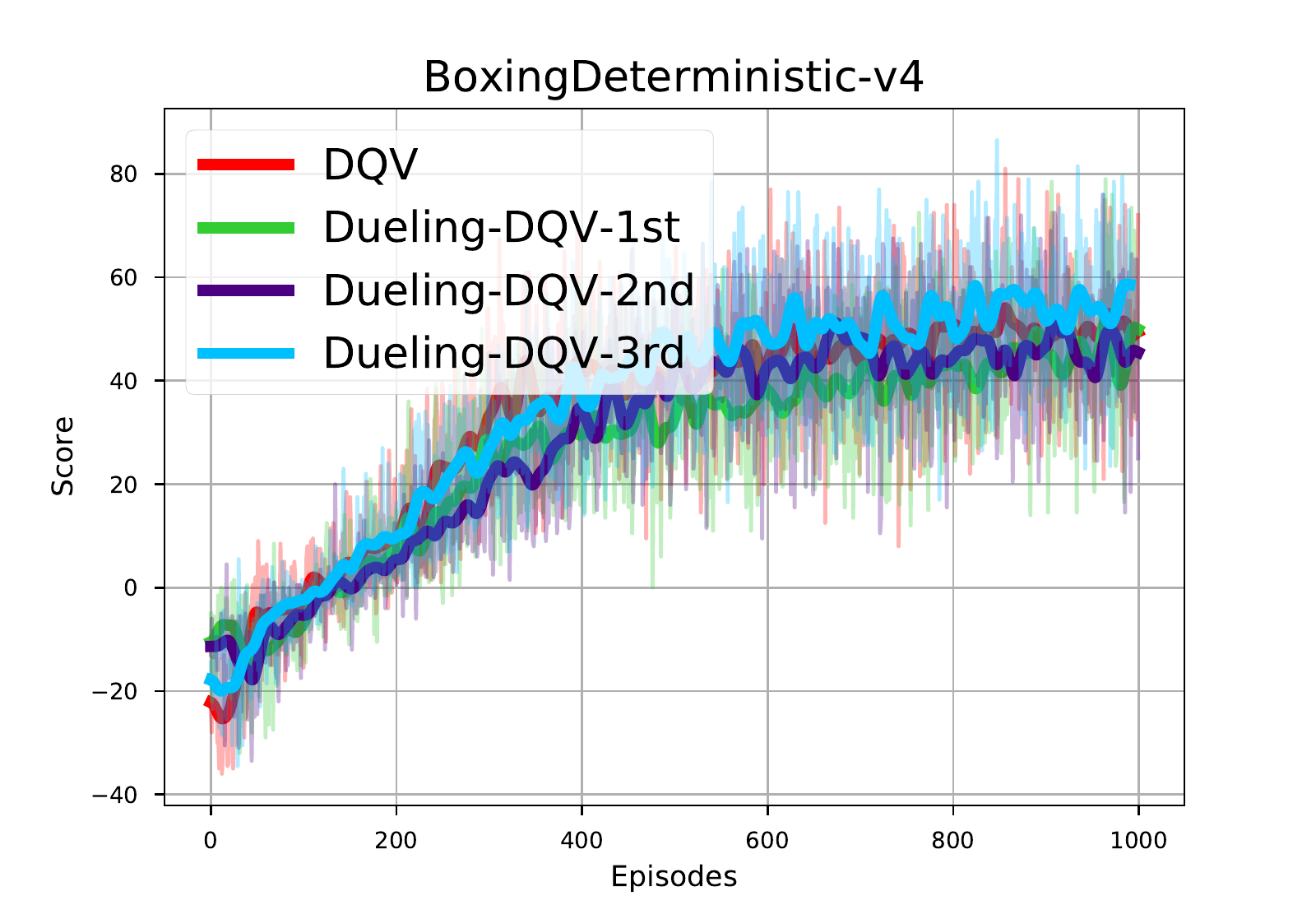}\hfill
\includegraphics[width=.3\textwidth]{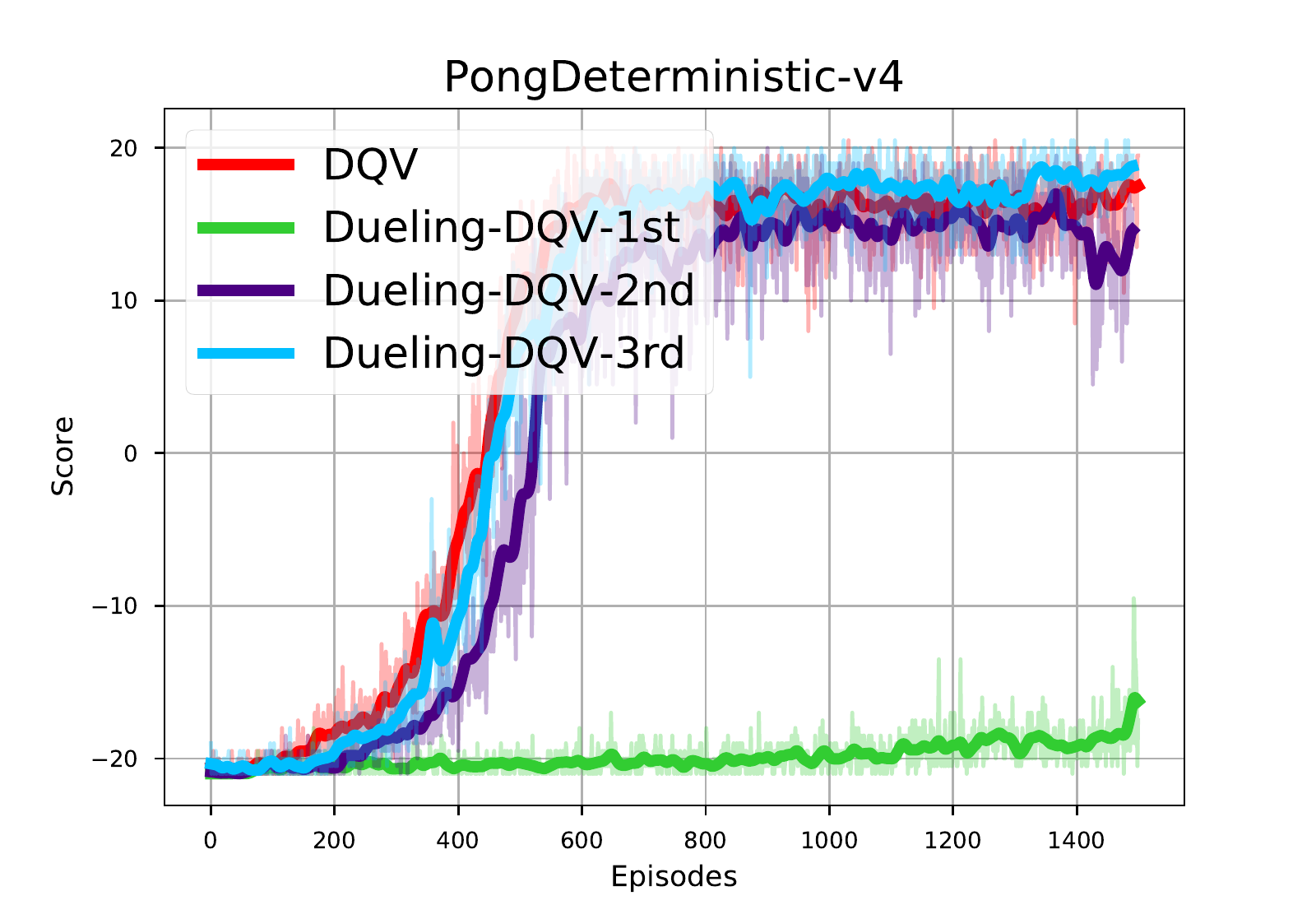}\hfill
\includegraphics[width=.3\textwidth]{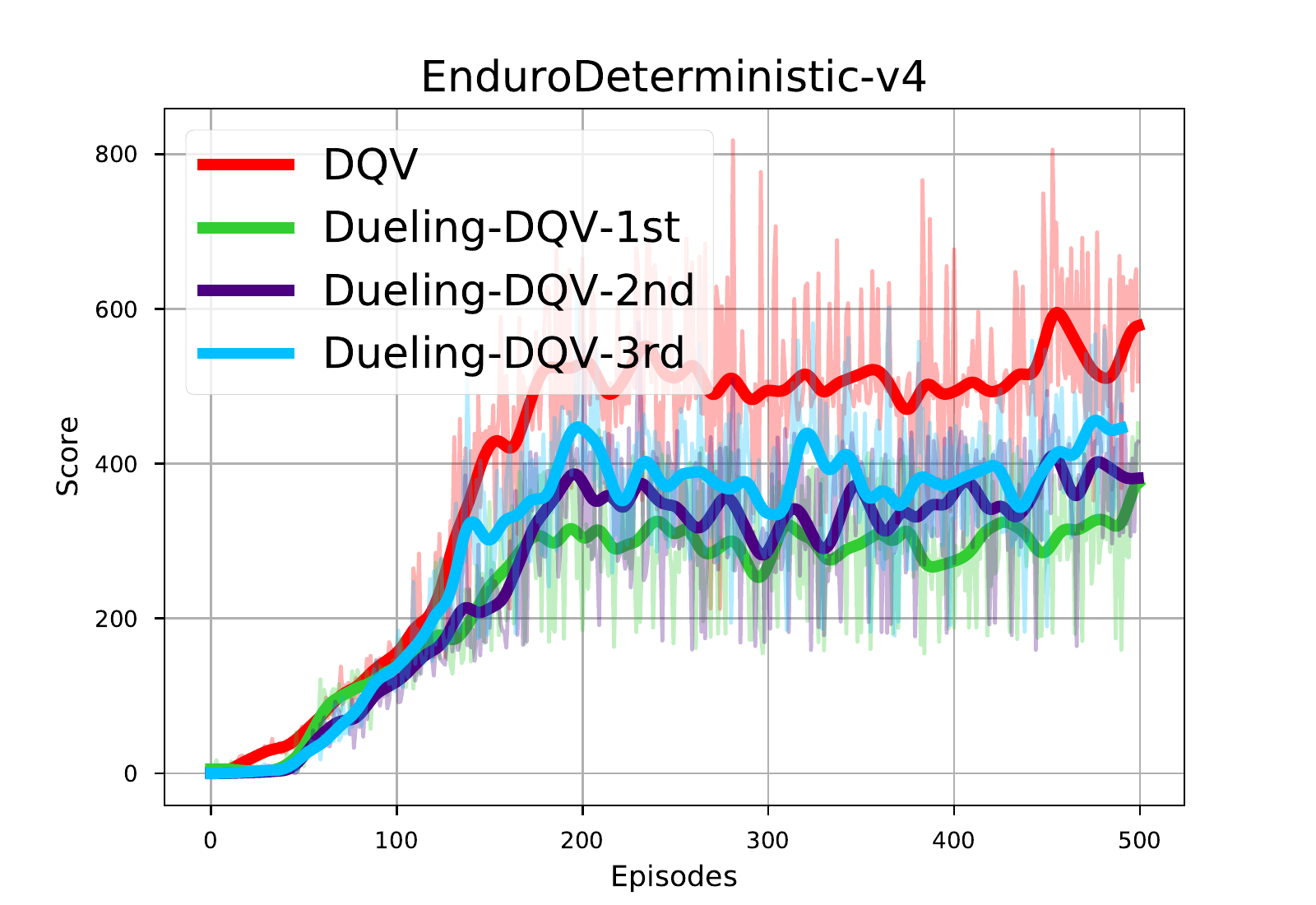}
\caption{The results obtained by the different Dueling-DQV architectures. On the \texttt{Boxing} and \texttt{Pong} environments similar results as the ones obtained by DQV can be achieved. However on the \texttt{Enduro} environment none of the proposed architectures is able to perform as well as DQV. This suggests that the most beneficial strategy for approximating two value functions is with two distinct neural networks.}
\label{fig:dueling-dqv}

\end{figure}

Better results have been obtained with a neural architecture which uses a specific hidden layer before the outputs of the $V$ and $Q$ functions (Fig. \ref{fig:dueling_dqv}). We refer to this algorithm as Dueling-DQV, where Dueling-DQV-1st corresponds to an architecture in which the $V$ function is learned after the first convolution block, Dueling-DQV-2nd after the second block and Dueling-DQV-3rd after the third. As reported in Fig. \ref{fig:dueling-dqv} on the simple \texttt{Boxing} environment we can observe that no matter where the state-value layer is positioned, all versions of Dueling-DQV perform as well as the original DQV algorithm. On the more complicated \texttt{Pong} environment, however, the performance of Dueling-DQV starts to be strongly affected by the depth of the state-value layer. The only case in which Dueling-DQV performs as well as DQV is when the $V$ function is learned after all three convolutional layers. More interestingly, on the most complicated \texttt{Enduro} environment, the performance of DQV cannot be matched by any of the Dueling architectures, in fact, the rewards obtained by Dueling-3rd and DQV differ with $\approx 200$ points. These results seem to suggest that the parameters of DQV can be reduced, as long as some specific value-layers are added to the neural architecture. However, this approach presents limitations. As shown by the results obtained on the \texttt{Enduro} environment, the best version of DQV remains the one in which two distinct neural networks approximate the $V$ and $Q$ functions.

\subsection{The Deep Quality-Value-Max Algorithm (Question 2)}
\label{sec:dqv_max}

\begin{figure}[ht!]

\includegraphics[width=.3\textwidth]{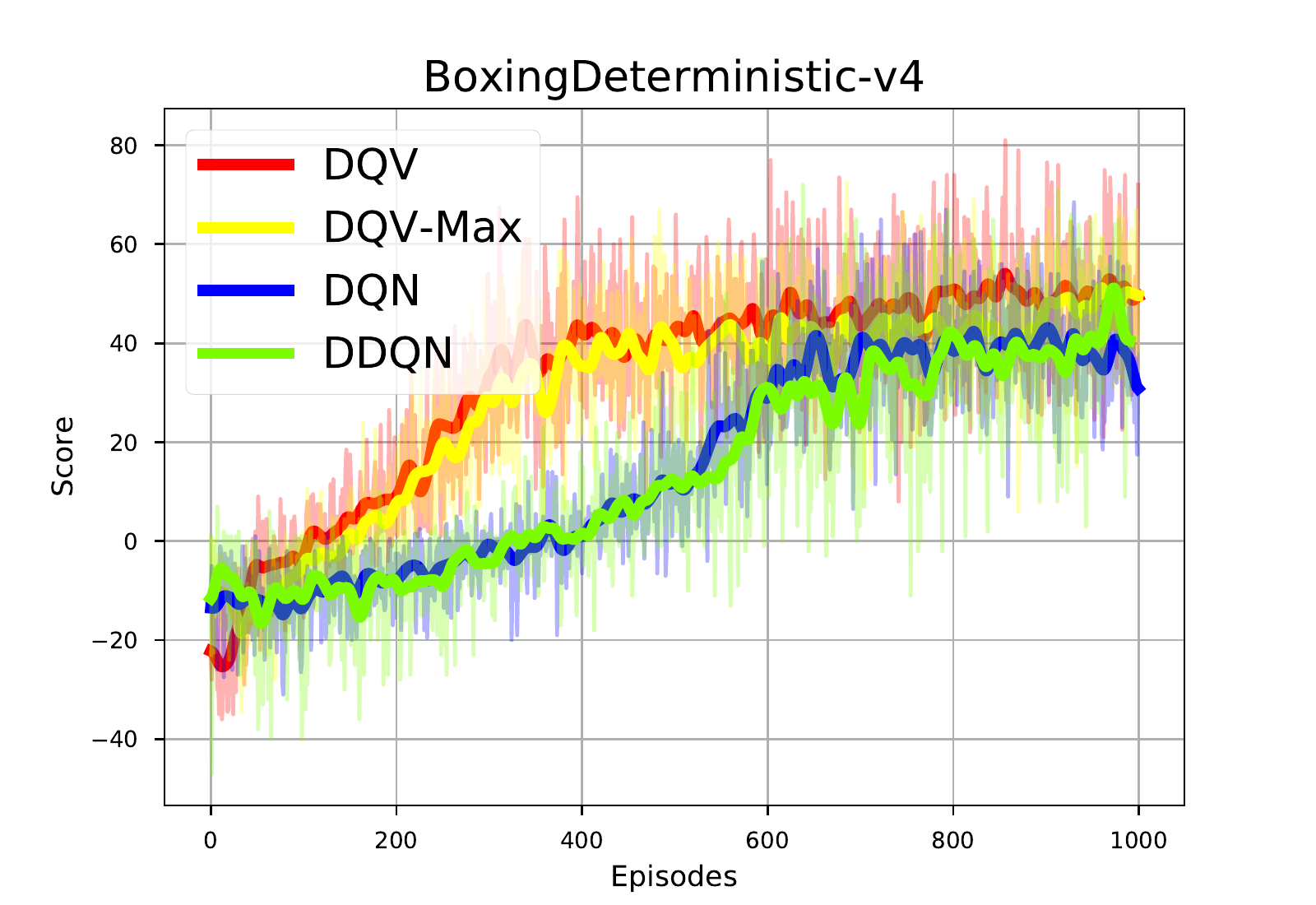}\hfill
\includegraphics[width=.3\textwidth]{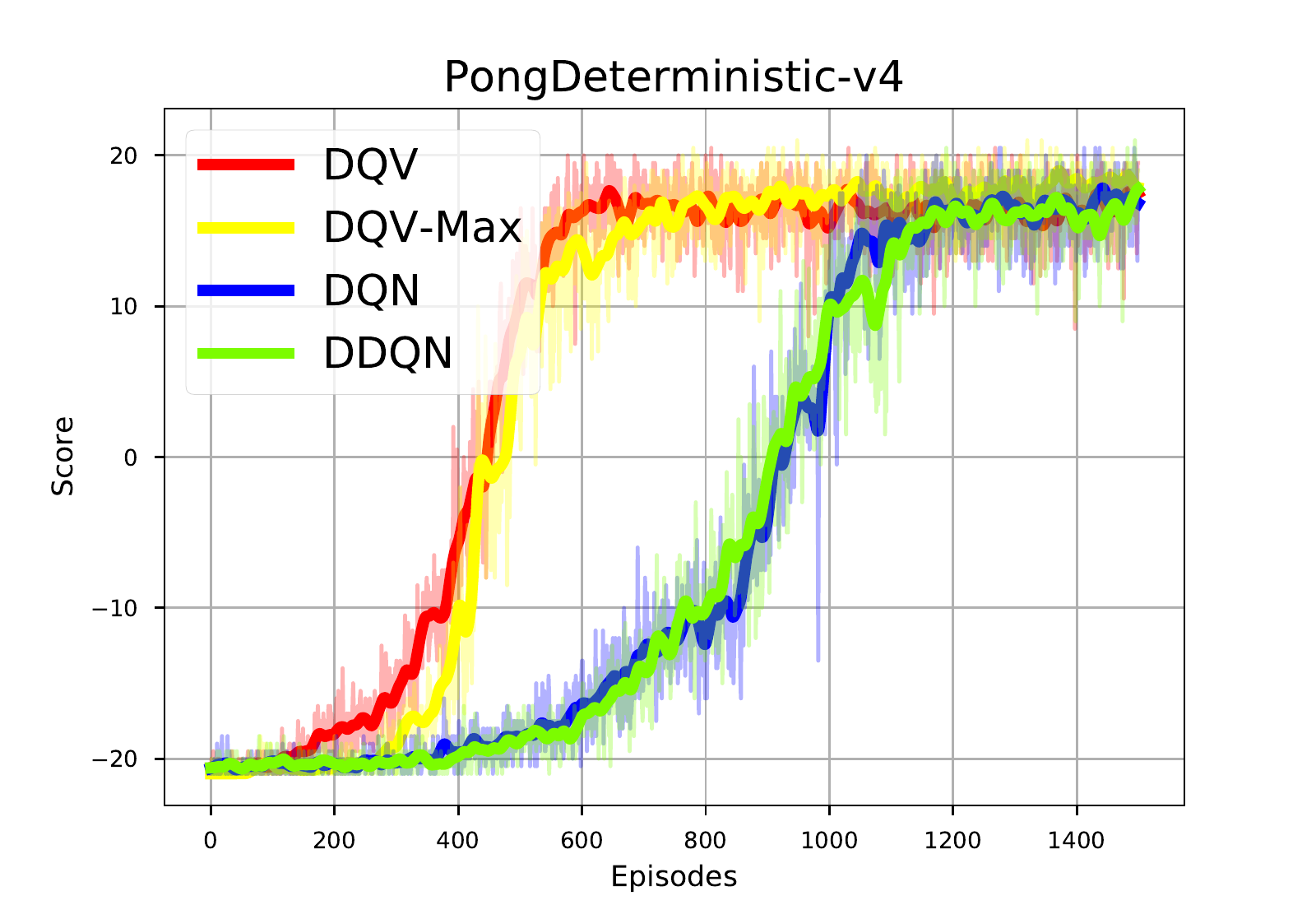}\hfill
\includegraphics[width=.3\textwidth]{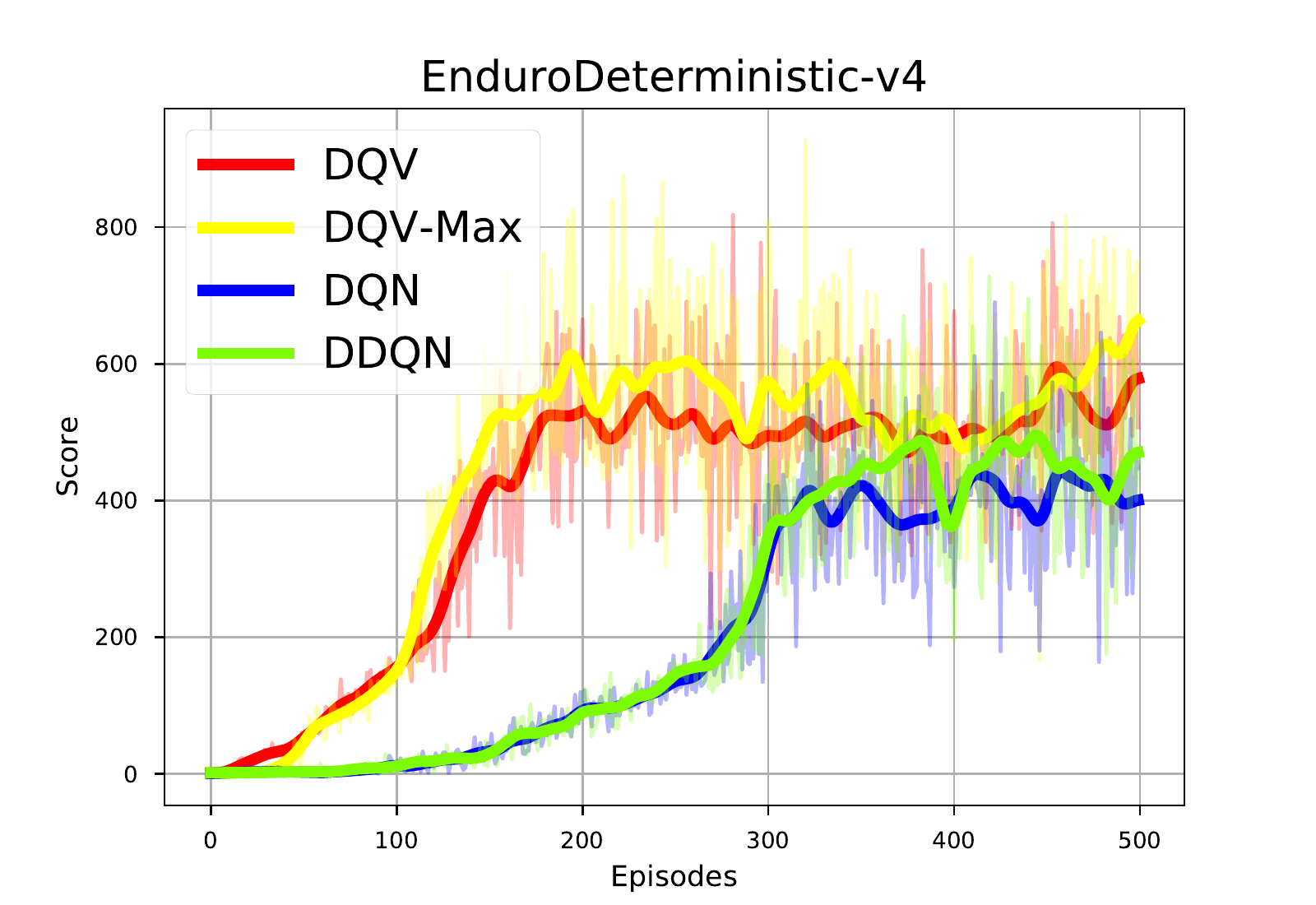}\caption{The results obtained by the DQV-Max algorithm on the \texttt{Boxing, Pong} and \texttt{Enduro} environments. DQV-Max is able to learn as fast and even better than DQV, suggesting that jointly approximating two value functions is also beneficial in an \textit{off-policy} learning setting.}
\label{fig:dqv-max-results}
\end{figure}

 By investigating the performance of the novel DQV-Max algorithm, promising results have been obtained. As we can see in Fig. \ref{fig:dqv-max-results}, DQV-Max has a comparable, and sometimes even better performance than the DQV algorithm, therefore strongly outperforming DQN and DDQN. It learns as fast when it comes to the \texttt{Boxing} and \texttt{Pong} environments and achieves an even higher cumulative reward on the \texttt{Enduro} environment, making it the overall best performing algorithm. These results remark the benefits of jointly learning an approximation of the $V$ function and the $Q$ function, and show that this approach is just as beneficial when it comes to an \textit{off-policy} learning setting than it is in an \textit{on-policy} learning one. We can therefore answer affirmatively to the second research question analyzed in this work.

\subsection{Reducing the overestimation bias of the $Q$ function (Question 3)}
\label{sec:divergence}

\begin{figure}[ht!]
  \makebox[\textwidth][c]{%
  \subfloat[]{\scalebox{0.4}{\includegraphics{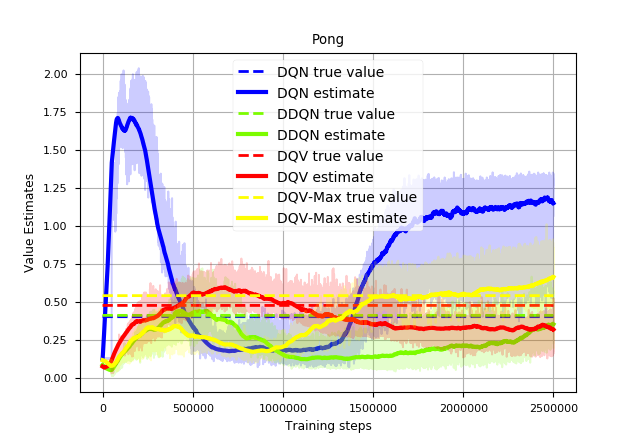}} \label{fig:overestimation_1}}
  \quad
  \subfloat[]{\scalebox{0.4}{\includegraphics{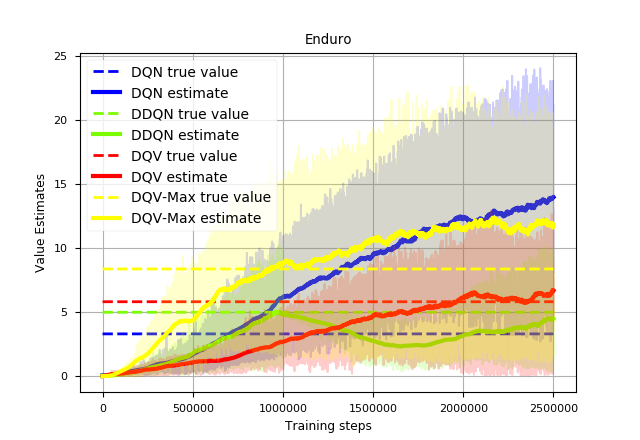}} \label{fig:overestimation_2}}
  }
\caption{A representation of the value estimates that are obtained by each algorithm at training time (denoted by the full lines), with respect to the actual discounted return obtained by an already trained agent (dashed lines). We can observe that on \texttt{Pong} DQN's estimates rapidly grow before getting closer to the baseline discounted value, while DQV and DQV-Max do not show this behavior. Their value estimates are more similar to the ones of the DDQN algorithm. This is especially the case for DQV. On the \texttt{Enduro} environment, DQN's estimates do not stop increasing over time and keep moving away from the actual values of each state while DQV's value estimates remain bounded and do not diverge as much. DQV-Max diverges more
than DQV but still not as much as DQN. DQN's behavior corresponds to the one observed in \cite{van2016deep} and can be linked to the overestimation bias of the algorithm, (which is fully corrected by DDQN). This suggests that DQV and DQV-Max might perform so well on the ALE environment because they are less prone to overestimate the $Q$ function. Furthermore, it is worth noting the different baseline values which denote the averaged discounted return estimates when it comes to the \texttt{Enduro} environment: the lines representing the true values of the final policy are very different among algorithms, indicating that DQV and DQV-Max do not only learn more accurate value estimates but also lead to better final policies.}
\end{figure}

To investigate whether DQV and DQV-Max suffer from the overestimation bias of the $Q$ function we have performed the following experiment. As proposed in \cite{van2016deep} we monitor the $Q$ function of the algorithms at training time by computing the averaged $\underset{a \in \cal A}{\max}\:Q(s_{t+1}, a)$ over a set ($n$) of full evaluation episodes as defined by $\frac{1}{n}\sum_{t=1}^{n}\underset{a \in \cal A}{\max}\:Q(s_{t+1}, a;\theta).$ We then compare these estimates with the averaged discounted return of all visited states which is given by the same agent that has already concluded training. This provides a reliable baseline for measuring whether the estimated $Q$ values diverge from the ones which should actually be predicted. We report these results in Fig. \ref{fig:overestimation_1} and \ref{fig:overestimation_2}, where for each plot the dashed line corresponds to the actual averaged discounted return of the visited states, while the full lines correspond to the value estimates that are computed by each algorithm.

Our results show that DQV and DQV-Max seem to suffer less from the overestimation bias of the $Q$ function since both algorithms learn more stable and accurate value estimates. This allows us to answer both hypotheses introduced in Sec. \ref{sec:deadly_triad} affirmatively: jointly approximating the $V$ function and the $Q$ function can prevent DRL from diverging, since this learning dynamic allows the algorithms to estimate more realistic and not increasingly large $Q$ values. However, this becomes harder to achieve once the algorithm learns \textit{off-policy}. By analyzing the plot presented in Fig. \ref{fig:overestimation_2}, we can observe that the value estimates of DQV-Max are still higher from the ones which should be predicted by the end of training. However, differently, from the ones of DQN, they get bounded over time, therefore resulting in a less strong divergence. It is also worth noting how in our experiments the DDQN algorithm fully corrects the overestimation bias of DQN as expected.

\begin{figure}[ht!]
  \makebox[\textwidth][c]{%
  \subfloat[]{\scalebox{0.4}{\includegraphics{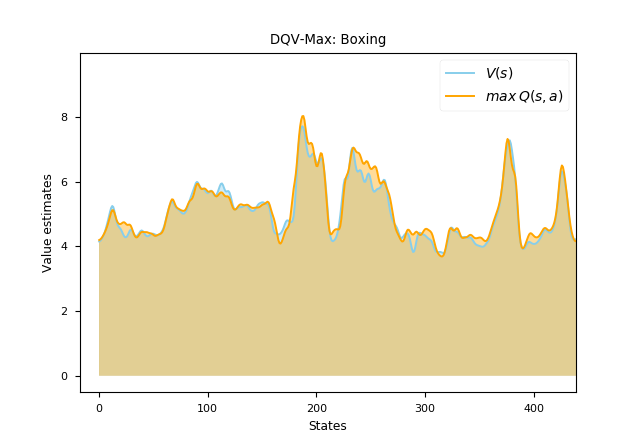}} \label{fig:value_estimates_1}}
  \quad
  \subfloat[]{\scalebox{0.4}{\includegraphics{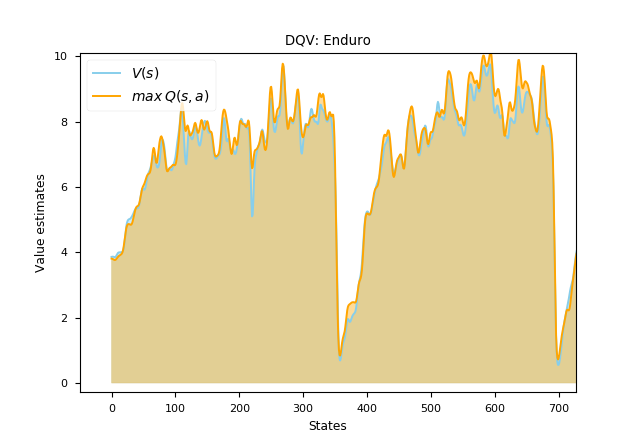}} \label{fig:value_estimates_2}}
  } \caption{The results showing that the value estimates of the $V$ network are never significantly higher than the ones coming from the $Q$ network for a set of randomly sampled states. On the contrary, the $V(s)$ is mostly lower than the respective $\max\:Q(s,a)$ estimate. This suggests that DQV and DQV-Max do not overestimate the state-value function instead of the state-action value function.}
\label{fig:sign-function}
\end{figure}

One could argue that DQV and DQV-Max might not be overestimating the $Q$ function because they are overestimating the $V$ function instead. In order to verify this, we have investigated whether the estimates of the state-value network are higher than the ones coming from the state-action value network. Similarly to the previous experiment we use the models that are obtained at the end of training and randomly sample a set of states from the ALE which are then used in order to compute $V(s)$ and $\max\:Q(s,a)$. We then investigate whether $V(s)$ is higher than the respective $\max\:Q(s,a)$ estimate. As can be seen in Fig. \ref{fig:sign-function} this is almost never the case, which empirically shows that both DQV and DQV-Max do not overestimate the $V$ function instead of the $Q$ function.

\section{Final Evaluation Performance}
\label{sec:final_performance}

In order to better evaluate the performance of all analyzed algorithms we now present results which have been obtained on a larger set of \texttt{Atari} games. The main aim of this section is to investigate whether the DQV-Max algorithm can be as successful as DQV, and to test both algorithms over a larger, more significant set of \texttt{Atari} games than the ones which were used when the DQV-Learning algorithm was initially introduced \cite{sabatelli2018deep}. Our evaluation protocol is based on the widely used \texttt{no-op action} evaluation regime \cite{mnih2015human,van2016deep} in which the policies of the agents are tested over a series of episodes that all start by executing a set of partially random actions. It is known that this is an effective way of testing the level of generalization of the learned policies. In Table \ref{tab:results} we report the baseline scores that are obtained by random-play and by an expert-human player which come from \cite{mnih2015human}. These scores are necessary for evaluating whether our DRL algorithms are able to achieve the super-human performance on the \texttt{Atari} benchmark which has made the DQN algorithm so popular. Furthermore, in order to test whether the algorithms of the DQV family can be considered as two valid novel alternatives within synchronous model-free DRL, we again compare their performance with DQN and DDQN. Our results, reported in Table \ref{tab:results}, show the best performing algorithm in a green cell, while the second-best performing agent is shown in a yellow cell. When the scores over games are equivalent, like on the \texttt{Pong} environment, we report in the green and yellow cells the fastest and second fastest algorithm with respect to its convergence time.

Overall, we can observe that both DQV and DQV-Max successfully master all the different games on which they have been tested on, with the exception for the \texttt{Montezuma's Revenge} game, which will probably require more sophisticated exploration strategies \cite{fortunato2017noisy} to be mastered. It is worth noting that the novel DQV-Max algorithm introduced in this work is the overall best performing algorithm, and that the algorithms of the DQV-family are the only ones which are able to obtain super-human performance on the games \texttt{Bank Heist} and \texttt{Enduro}. Lastly, even if the performance of DQV and DQV-Max is close to the one of DQN and DDQN it is worth mentioning that, as highlighted by the learning curves presented in Sec. \ref{sec:dqv_max}, these algorithms converge significantly faster than DQN and DDQN, therefore resulting in faster training.

\begin{table*}[ht]
\tiny
\caption{The results obtained by DQV and DQV-Max on a larger set of \texttt{Atari} games when following the \texttt{no-op action} evaluation regime used in \cite{mnih2015human} and \cite{van2016deep}. We can observe that both algorithms are overall able to perform equally well as DQN and DDQN, with the main difference that, as reported in Fig. \ref{fig:dqv-max-results}, these algorithms converge significantly faster.}
\centering
\begin{tabular}{lrrrrrr}
\hline
Environment & Random & Human & DQN\cite{mnih2015human} & DDQN\cite{van2016deep} & DQV & DQV-Max\\
\hline
\texttt{Asteroids} &719.10 &13156.70 &\cellcolor{yellow!25}1629.33 &930.60 &1445.40 & \cellcolor{green!25}1846.08\\
\texttt{Bank Heist} &14.20 & 734.40 & 429.67 & 728.30 & \cellcolor{green!25}1236.50 & \cellcolor{yellow!25}1118.28 \\
\texttt{Boxing} &0.10 & 4.30 & 71.83 & \cellcolor{green!25}81.70 & 78.66 & \cellcolor{yellow!25}{80.15} \\
\texttt{Crazy Climber} &10780.50 & 35410.50 & \cellcolor{green!25}114103.33 & 101874.00 & \cellcolor{yellow!25}108600.00 & 1000131.00\\
\texttt{Enduro} &0.00 & 309.60 & 301.77 & 319.50 & \cellcolor{yellow!25}829.33 & \cellcolor{green!25}875.64 \\
\texttt{Fishing Derby} &-91.70 & 5.50 & -0.80 & \cellcolor{yellow!25}20.30 & 1.12 & \cellcolor{green!25}20.42  \\
\texttt{Frostbite} &65.20 & 4334.70 & \cellcolor{green!25}328.33 & 241.50 & 271.86 & \cellcolor{yellow!25}281.36 \\
\texttt{Ice Hockey} &-11.20 & 0.90 & \cellcolor{yellow!25}-1.60 & -2.40 & -1.88 & \cellcolor{green!25}-1.12\\
\texttt{James Bond} &29.00 & 406.70 & \cellcolor{green!25}{576.67} & \cellcolor{yellow!25}{438.00} & 372.41 &375.00 \\
\texttt{Montezuma's Revenge} &0.00 & 4366.70 & 0.00 & 0.00 & 0.00 & 0.00\\
\texttt{Ms.Pacman} &307.30 & 15693.40 & 2311.00 & 3210.00 & \cellcolor{green!25}3590.00 & \cellcolor{yellow!25}3390.00\\
\texttt{Pong} &-20.70 &9.30 & 18.90 & 21.00 & \cellcolor{yellow!25}21.00 & \cellcolor{green!25}21.00\\
\texttt{Road Runner} &11.50 &7845.00 &18256.67  &\cellcolor{green!25}48377.00  &\cellcolor{yellow!25}39290.00  & 20700.00\\
\texttt{Zaxxon} &32.50 &9173.30 &4976.67  &\cellcolor{yellow!25}10182.00  &\cellcolor{green!25}10950.00  & 8487.00\\

\hline
\end{tabular}
\label{tab:results}
\end{table*}

\section{Conclusion}

In this work, we have made one step towards properly characterizing a new family of DRL algorithms which simultaneously learns an approximation of the $V$ function and the $Q$ function. We have started by thoroughly analyzing the DQV algorithm \cite{sabatelli2018deep}, and have shown in Sec. \ref{sec:shared_dqv} that one key component of DQV is to use two independently parameterized neural networks for learning the $V$ and $Q$ functions. We have then borrowed some ideas from DQV to construct a novel DRL algorithm in Sec. \ref{sec:dqv_max} in order to show that approximating two value functions instead of one is just as beneficial in an \textit{off-policy} learning setting as it is in an \textit{on-policy} learning one. We have then studied how the DQV and DQV-Max algorithms are related to the DRL \textit{Deadly Triad}, and hypothesized that the promising results obtained by both algorithms, could partially be achieved because both algorithms could suffer less from the overestimation bias of the $Q$ function. From Sec. \ref{sec:divergence} we have concluded that this was indeed the case, even though DQV-Max seems to be more prone to suffer from this phenomenon.
We have then ended the paper with an in-depth empirical analysis of the studied algorithms in Sec. \ref{sec:final_performance}, which generalizes all the results used in the previous sections to a larger set of DRL test-beds. To conclude, this paper sheds some light on the benefits that could come from approximating two value functions instead of one, and properly characterizes a new family of DRL algorithms which follow such learning dynamics.

\bibliographystyle{plain}
\bibliography{my_bib}

\begin{thebibliography}{10}

\bibitem{achiam2019towards}
Joshua Achiam, Ethan Knight, and Pieter Abbeel.
\newblock Towards characterizing divergence in deep {Q}-learning.
\newblock {\em arXiv preprint arXiv:1903.08894}, 2019.

\bibitem{bellemare2013arcade}
Marc~G Bellemare, Yavar Naddaf, Joel Veness, and Michael Bowling.
\newblock The arcade learning environment: An evaluation platform for general
  agents.
\newblock {\em Journal of Artificial Intelligence Research}, 47:253--279, 2013.

\bibitem{bellman1966dynamic}
Richard Bellman.
\newblock Dynamic programming.
\newblock {\em Science}, 153(3731):34--37, 1966.

\bibitem{boyan1995generalization}
Justin~A Boyan and Andrew~W Moore.
\newblock Generalization in reinforcement learning: Safely approximating the
  value function.
\newblock In {\em Advances in neural information processing systems}, pages
  369--376, 1995.

\bibitem{caruana1997multitask}
Rich Caruana.
\newblock Multitask learning.
\newblock {\em Machine learning}, 28(1):41--75, 1997.

\bibitem{fortunato2017noisy}
Meire Fortunato, Mohammad~Gheshlaghi Azar, Bilal Piot, Jacob Menick, Ian
  Osband, Alex Graves, Vlad Mnih, Remi Munos, Demis Hassabis, Olivier Pietquin,
  et~al.
\newblock Noisy networks for exploration.
\newblock {\em arXiv preprint arXiv:1706.10295}, 2017.

\bibitem{fujimoto2018addressing}
Scott Fujimoto, Herke Hoof, and David Meger.
\newblock Addressing function approximation error in actor-critic methods.
\newblock In {\em International Conference on Machine Learning}, pages
  1582--1591, 2018.

\bibitem{hasselt2010double}
Hado~Van Hasselt.
\newblock Double {Q}-learning.
\newblock In {\em Advances in Neural Information Processing Systems}, pages
  2613--2621, 2010.

\bibitem{henderson2018deep}
Peter Henderson, Riashat Islam, Philip Bachman, Joelle Pineau, Doina Precup,
  and David Meger.
\newblock Deep reinforcement learning that matters.
\newblock In {\em Thirty-Second AAAI Conference on Artificial Intelligence},
  2018.

\bibitem{hessel2018rainbow}
Matteo Hessel, Joseph Modayil, Hado Van~Hasselt, Tom Schaul, Georg Ostrovski,
  Will Dabney, Dan Horgan, Bilal Piot, Mohammad Azar, and David Silver.
\newblock Rainbow: Combining improvements in deep reinforcement learning.
\newblock In {\em Thirty-Second AAAI Conference on Artificial Intelligence},
  2018.

\bibitem{lecun2015deep}
Yann LeCun, Yoshua Bengio, and Geoffrey Hinton.
\newblock Deep learning.
\newblock {\em Nature}, 521(7553):436, 2015.

\bibitem{lin1993reinforcement}
Long-Ji Lin.
\newblock Reinforcement learning for robots using neural networks.
\newblock Technical report, Carnegie-Mellon Univ Pittsburgh PA School of
  Computer Science, 1993.

\bibitem{mnih2015human}
Volodymyr Mnih, Koray Kavukcuoglu, David Silver, Andrei~A Rusu, Joel Veness,
  Marc~G Bellemare, Alex Graves, Martin Riedmiller, Andreas~K Fidjeland, Georg
  Ostrovski, et~al.
\newblock Human-level control through deep reinforcement learning.
\newblock {\em Nature}, 518(7540):529, 2015.

\bibitem{pong2018temporal}
Vitchyr Pong, Shixiang Gu, Murtaza Dalal, and Sergey Levine.
\newblock Temporal difference models: Model-free deep {RL} for model-based
  control.
\newblock {\em arXiv preprint arXiv:1802.09081}, 2018.

\bibitem{rummery1994line}
Gavin~A Rummery and Mahesan Niranjan.
\newblock {\em On-line Q-learning using connectionist systems}, volume~37.
\newblock University of Cambridge, Department of Engineering Cambridge,
  England, 1994.

\bibitem{sabatelli2018deep}
Matthia Sabatelli, Gilles Louppe, Pierre Geurts, and Marco Wiering.
\newblock Deep quality-value (dqv) learning.
\newblock In {\em Advances in Neural Information Processing Systems, Deep
  Reinforcement Learning Workshop}. Montreal, 2018.

\bibitem{schmidhuber2015deep}
J{\"u}rgen Schmidhuber.
\newblock Deep learning in neural networks: An overview.
\newblock {\em Neural networks}, 61:85--117, 2015.

\bibitem{sutton1988learning}
Richard~S Sutton.
\newblock Learning to predict by the methods of temporal differences.
\newblock {\em Machine learning}, 3(1):9--44, 1988.

\bibitem{sutton2018reinforcement}
Richard~S Sutton and Andrew~G Barto.
\newblock {\em Reinforcement learning: An introduction}.
\newblock MIT press, 2018.

\bibitem{tieleman2012lecture}
Tijmen Tieleman and Geoffrey Hinton.
\newblock Lecture 6.5-rmsprop: Divide the gradient by a running average of its
  recent magnitude.
\newblock {\em COURSERA: Neural networks for machine learning}, 4(2):26--31,
  2012.

\bibitem{van2018deep}
Hado Van~Hasselt, Yotam Doron, Florian Strub, Matteo Hessel, Nicolas Sonnerat,
  and Joseph Modayil.
\newblock Deep reinforcement learning and the deadly triad.
\newblock {\em arXiv preprint arXiv:1812.02648}, 2018.

\bibitem{van2016deep}
Hado Van~Hasselt, Arthur Guez, and David Silver.
\newblock Deep reinforcement learning with double {Q}-learning.
\newblock In {\em Thirtieth AAAI Conference on Artificial Intelligence}, 2016.

\bibitem{wang2016dueling}
Ziyu Wang, Tom Schaul, Matteo Hessel, Hado Van~Hasselt, Marc Lanctot, and Nando
  Freitas.
\newblock Dueling network architectures for deep reinforcement learning.
\newblock In {\em International Conference on Machine Learning}, pages
  1995--2003, 2016.

\bibitem{watkins1992q}
Christopher~JCH Watkins and Peter Dayan.
\newblock Q-learning.
\newblock {\em Machine learning}, 8(3-4):279--292, 1992.

\bibitem{wiering2005qv}
Marco~A Wiering.
\newblock {QV} (lambda)-learning: A new on-policy reinforcement learning
  algorithm.
\newblock In {\em Proceedings of the 7th European Workshop on Reinforcement
  Learning}, pages 17--18, 2005.

\bibitem{wiering2009qv}
Marco~A Wiering and Hado Van~Hasselt.
\newblock The {QV} family compared to other reinforcement learning algorithms.
\newblock In {\em Adaptive Dynamic Programming and Reinforcement Learning,
  2009. ADPRL'09. IEEE Symposium on}, pages 101--108. IEEE, 2009.

\end{thebibliography}

\section{Supplementary Material}

\subsection{Experimental Setup}
\label{sec:experiments}

 We follow the experimental setup which is widely used in the literature \cite{mnih2015human, van2016deep, hessel2018rainbow}. A convolutional neural network with three convolutional layers is used \cite{mnih2015human}, followed by a fully connected layer. The network receives the frames of the different games as input. When it comes to DQV and DQV-Max we use the same architecture for approximating the $V$ function and the $Q$ function with the only difference being the size of the output layer. For the $V$ network this is simply $1$, while for the $Q$ network there are as many output nodes as possible actions. Both networks are trained with the RMSprop optimizer \cite{tieleman2012lecture} initialized as in \cite{mnih2015human}. The version on the ALE environment corresponds to the \verb#Deterministic-v4# one which uses `Frame-Skipping', a design choice which lets the agent select an action every $4$th frame instead of every single one. Furthermore, we use the standard \texttt{Atari}-preprocessing scheme for resizing each frame of the environment to an $84 \times 84$ gray-scaled matrix and as exploration strategy we use for all algorithms the epsilon-greedy approach. Regarding the target networks which are used in all our experiments, we update their weights each time the agent has performed a total of 10,000 actions. Lastly, the discount factor $\gamma$ is set to 0.99 and the size of the memory buffer is set to contain 1,000,000 transitions. So far, we have run experiments on 14 different games for which each algorithm is tested by running $5$ different simulations with $5$ different random seeds. The code for reproducing our experiments can be found at \url{https://github.com/paintception/Deep-Quality-Value-Family-/tree/master/src}. We are currently testing our algorithms on as many games of the \texttt{Atari} benchmark as possible.

\subsection{Pseudocode of DQV and DQV-Max learning.}
This section reports the pseudo-codes of Deep Quality-Value (DQV) learning (Algorithm \ref{alg: dqv_algorithm}) and its novel Deep Quality-Value-Max (DQV-Max) extension (Algorithm \ref{alg: dqv_max_algorithm}). Each algorithm requires the initialization of two neural networks that are required for approximating the state-value function $\Phi$ and the state-action value function $\theta$. DQV requires a target network for estimating the state-value ($V$) function which is initialized as $\Phi^{-}$, whereas DQV-Max requires it for learning the state-action value ($Q$) function, therefore it is defined as $\theta^{-}$. For our experiments the Experience Replay buffer $D$ is set to contain at most 400,000 trajectories ($N$), from which we start sampling as soon as it contains 50000 ($\mathcal{N}$) trajectories ($\langle$ $s_{t}$, $a_{t}$, $r_{t}$, $s_{t+1}$ $\rangle$). $D$ is handled as a queue: when its maximum size is reached, every new sample stored in the queue overwrites the oldest one. We also initialize the counter-variable, total\_a which is required for keeping track of how many actions the agent has performed over time. Once it corresponds to the hyperparameter $c$ we update the weights of the target network with the ones of the main network. Regarding the targets that are constructed in order to compute the Temporal-Difference errors we refer to them as $y_t$ in the DQV algorithm, while as $v_t$ and $q_t$ when it comes to DQV-Max. This is done in order to highlight the fact that the latter algorithm requires the computation of two different targets to bootstrap from.


\begin{algorithm}[ht]
\caption{DQV Learning}\label{alg: dqv_algorithm}
\begin{algorithmic}[1]
\Require{Experience Replay Queue $D$ of maximum size $N$}{}
\Require{$Q$ network with parameters $\theta$}{}
\Require{$V$ networks with parameters $\Phi$ and $\Phi^{-}$} {}
\Require{total\_a = 0}
\Require{total\_e = 0}
\Require{c = $10000$}
\Require{$\mathcal{N} = 50000$}

\While{True}
\State{set $s_t$ as the initial state}
\While{$s_t$ is not over}
    \State{select $a_t\in\mathcal{A}$ for $s_{t}$ with policy $\pi$ (using the epsilon-greedy strategy)}
    \State{get $r_{t}$ and $s_{t+1}$}
    \State{store $\langle s_{t}, a_{t}, r_{t}, s_{t+1}\rangle$ in $D$}
    \State{$s_t := s_{t+1}$}
    \State{total\_e += 1}
    \If{total\_e $\geqq \mathcal{N}$}
        \State{sample a minibatch $B=\{\langle s^i_t, a_t^i, r^i_t, s^i_{t+1}\rangle|i=1,\ldots,32\}$ of size 32 from $D$}
        \For{i = 1 to 32}
        \If{$s^i_{t+1}$ is over}
            \State{$y^i_{t} := r^i_{t}$}
        \Else
            \State{$y^i_{t} := r^i_{t} + \gamma V(s^i_{t+1}, \Phi^{-})$}
         \EndIf
         \EndFor
         \State{$\theta := \underset{\theta}{\argmin} \sum_{i=1}^{32} (y^i_{t} - Q(s^i_{t}, a^i_{t}, \theta))^{2}$}
        \State{$\Phi := \underset{\Phi}{\argmin} \sum_{i=1}^{32}(y^i_{t} - V(s^i_{t}, \Phi))^{2}$}
        \State{total\_a += 1}
        \If{total\_a = $c$}
            \State{$\Phi^{-}$ := $\Phi$}
            \State{total\_a := 0}
         \EndIf
    \EndIf
\EndWhile
\EndWhile

\end{algorithmic}
\end{algorithm}



\begin{algorithm}[ht]
\caption{DQV-Max Learning}\label{alg: dqv_max_algorithm}
\begin{algorithmic}[1]
\Require{Experience Replay Queue $D$ of maximum size $N$}{}
\Require{$Q$ networks with parameters $\theta$ and $\theta^{-}$}{}
\Require{$V$ network with parameters $\Phi$} {}
\Require{total\_a = 0}
\Require{total\_e = 0}
\Require{c = $10000$}
\Require{$\mathcal{N} = 50000$}

\While{True}
\State{set $s_t$ as the initial state}
\While{$s_t$ is not over}
    \State{select $a_t\in\mathcal{A}$ for $s_{t}$ with policy $\pi$ (using the epsilon-greedy strategy)}
    \State{get $r_{t}$ and $s_{t+1}$}
    \State{store $\langle s_{t}, a_{t}, r_{t}, s_{t+1}\rangle$ in $D$}
    \State{$s_t := s_{t+1}$}
    \State{total\_e += 1}
    \If{total\_e $\geqq \mathcal{N}$}
            \State{sample a minibatch $B=\{\langle s^i_t, a_t^i, r^i_t, s^i_{t+1}\rangle|i=1,\ldots,32\}$ of size 32 from $D$}
        \For{i = 1 to 32}
        \If{$s^i_{t+1}$ is over}
            \State{$v^i_{t} := r^i_{t}$}
            \State{$q^i_{t} := r^i_{t}$}
        \Else
            \State{$v^i_{t} := r^i_{t} + \gamma \: \underset{a\in \mathcal{A}}{\max}\: (Q(s^i_{t+1}, a, \theta^{-}))$}
            \State{$q^i_{t} := r^i_{t} + \gamma \: V(s^i_{t+1}, \Phi)$}
         \EndIf
         \EndFor
         \State{$\theta := \underset{\theta}{\argmin}\: \sum_{i=1}^{32} (q^i_{t} - Q(s^i_{t}, a^i_{t}, \theta))^{2}$}
        \State{$\Phi := \underset{\Phi}{\argmin}\: \sum_{i=1}^{32} (v^i_{t} - V(s^i_{t}, \Phi))^{2}$}
        \State{total\_a += 1}
        \If{total\_a = $c$}
            \State{$\theta^{-}$ := $\theta$}
            \State{total\_a := 0}
         \EndIf
    \EndIf
\EndWhile
\EndWhile

\end{algorithmic}
\end{algorithm}

\end{document}